\documentclass{article} 
\usepackage{arxiv}
\usepackage[utf8]{inputenc} 
\usepackage[T1]{fontenc}    
\usepackage{hyperref}       
\usepackage{url}            
\usepackage{times}
\usepackage{graphicx}
\usepackage{amssymb}
\usepackage{url,hyperref}
\usepackage{multirow}
\usepackage{comment} 
\usepackage{graphics}
\usepackage{caption}
\usepackage{subcaption}
\usepackage{color}
\usepackage{amsmath}
\usepackage{booktabs}
\usepackage{mdframed}
\usepackage{algpseudocode}
\usepackage{algorithm}
\newtheorem{theorem}{Theorem}[section]
\newtheorem{corollary}{Corollary}[theorem]

\newtheorem{prop}{Proposition}
\title{EcoVal: An Efficient Data Valuation Framework for Machine Learning}
\date{}
\author{
  Ayush K Tarun$^*$ \\
  RespAI Lab, India\\
  \texttt{ayushtarun210@gmail.com} \\
  \And
  Vikram S Chundawat$^*$ \\
  RespAI Lab, India\\
  \texttt{vikram2000b@gmail.com} \\
  \And
  Murari Mandal \textdagger \\
  RespAI Lab, KIIT Bhubaneswar, India\\
  \texttt{murari.mandalfcs@kiit.ac.in} \\ 
 \And
  Hong Ming Tan\\
  NUS Business School\\
  National University of Singapore\\
  \texttt{thm@nus.edu.sg} \\
 \And
  Bowei Chen \\
  Adam Smith Business School\\
  University of Glasgow\\
  \texttt{bowei.chen@glasgow.ac.uk} \\
 \And
  Mohan Kankanhalli\\
  School of Computing\\
  National University of Singapore\\
  \texttt{mohan@comp.nus.edu.sg} \\
}

\begin{document}
\maketitle
\def\thefootnote{*}\footnotetext{These authors contributed equally to this work}
\def\thefootnote{\textdagger}
\footnotetext{Corresponding author}

\begin{abstract}
Quantifying the value of data within a machine learning workflow can play a pivotal role in making more strategic decisions in machine learning initiatives. The existing Shapley value based frameworks for data valuation in machine learning are computationally expensive as they require considerable amount of repeated training of the model to obtain the Shapley value. In this paper, we introduce an efficient data valuation framework EcoVal, to estimate the value of data for machine learning models in a fast and practical manner. Instead of directly working with individual data sample, we determine the value of a cluster of similar data points. This value is further propagated amongst all the member cluster points. We show that the overall value of the data can be determined by estimating the intrinsic and extrinsic value of each data. This is enabled by formulating the performance of a model as a \textit{production function}, a concept which is popularly used to estimate the amount of output based on factors like labor and capital in a traditional free economic market. We provide a formal proof of our valuation technique and elucidate the principles and mechanisms that enable its accelerated performance. We demonstrate the real-world applicability of our method by showcasing its effectiveness for both in-distribution and out-of-sample data. This work addresses one of the core challenges of efficient data valuation at scale in machine learning models. The code is available at \underline{https://github.com/respai-lab/ecoval}.
\end{abstract}


\section{Introduction}
Data valuation is a pivotal concern in modern machine learning (ML) and data analytics, where the quality and worth of data have profound implications for decision-making, model performance, and data marketplace. 
Quantifying the worth of data plays an important role in data pricing and regulation compliance~\cite{wadhwa2020economic,survey_datapricing}, removing low-value/noisy data from the training set~\cite{tang2021data,karlavs2022data}, and incentivizing data sharing by personal data monetization~\cite{jia2019towards,ghorbani2019data,ghorbani2020distributional,ilyas2022datamodels}. In a ML framework, the quality of data determines the effectiveness of the final model. Therefore, identifying high and low value data through data valuation would yield significant benefits for a wide range of machine learning applications.\par

\textbf{Background:} In recent studies, a cooperative game theory concept, Shapley value~\cite{shapley201617} has been frequently used for data valuation in supervised ML~\cite{ghorbani2019data,jia2019towards,ghorbani2020distributional}. It offers a desirable property of equitable reward allocation. The data Shapley and its extensions~\cite{ghorbani2019data,ghorbani2020distributional,kwon2021efficient,kwon2022beta} have empirically shown the effectiveness of Shapley value based valuation in a fixed dataset as well as in a particular distribution of data, allowing for out-of-time data valuation. The value of a data point in ML relies on its individual contribution to the model's performance and its relationship with other data points utilized during training. The presence of similar data in the training set can dilute the significance of individual points. To account for these interactions, data Shapley methods evaluate the contribution of each point by determining how its absence affects the overall performance of the model. This process usually involves repeatedly training the model with the selective exclusion of certain instances or subsets, thereby identifying those with the most substantial impact. The impact is measured by the observing the change in the performance score of the ML model. However, this incurs a high computational cost, typically requiring model training runs in the order of $O(n^{2})$ in current methodologies, where $n$ is the total number of data points in the dataset.\par

\textbf{Motivation:} While offering insightful analyses of data point significance and alleviating the issue of poor discrimination of data quality in leave-one-out (LOO) error methods, existing data Shapley based frameworks~\cite{ghorbani2019data,ghorbani2020distributional,wang2023data,kwon2022beta} suffer from a high computational cost. The need for a higher number of repeated training sessions for a model, as required by these methods, leads to inefficiencies in both time and resource utilization. Furthermore, this inefficiency translates to an increased carbon footprint due to the energy requirements of training, thereby exacerbating climate change concerns~\cite{lacoste2019quantifying}. The development of scalable algorithms capable of handling extensive datasets is essential for practical use of data valuation in real-world applications. 

\textbf{Our Contribution:} 
We adopt a two-step approach where the valuation is performed at cluster-level first and the value is further divided among the cluster members. The similar data points are represented through a cluster which significantly reduces the total number of data points to deal with during training phase of the valuation process. At cluster level, we can use a simple LOO error for valuation since there is minimal possibility (almost zero) of a similar datum to be found in other clusters. The difficulty however, is to divide the value at each cluster among the cluster members. To address this issue, a novel approach is proposed based on \textit{production functions} in economics. Our two-step approach aims to significantly speed up the valuation process in comparison to the Truncated Monte Carlo (TMC) Shapley.\par
In this paper, we introduce a a novel framework based on Leave Cluster Out (LCO) and production functions for data valuation in machine learning. The framework is computationally efficient, with theoretical and empirical verifications. The following are the key contributions of our work:\par
\textit{\textbf{Novel Framework:}} The intuition behind our framework is that we find a group of similar items and estimate this cluster's marginal contribution. As similar data items are bound to have similar values, we extend this principle to estimate cluster-level value  through Leave Cluster Out (LCO).\par
We introduce a \textit{production function} formulation representing the relation between the data and its utility in a model. We show that this formulation can be used to estimate the value of individual data based on the value of each cluster.\par 
\textit{\textbf{Computational Efficiency:}} We estimate the intrinsic and extrinsic value of each data point to determine the individual data value. By checking only the marginal contribution of the representative data point of a cluster, we substantially reduce the overhead of creating multiple subsets containing similar data points. Our approach is scalable to large datasets without being limited by the presence of similar data points in the dataset.

\textit{\textbf{Theoretical Proof:}} We provide a theoretical proof of our data valuation method. We also show that the valuation obtained by our method has negligible error margin when compared with the vanilla Shapley value approximation method.

\textit{\textbf{Empirical Evaluation:}} We conduct experiments with machine learning models on MNIST, CIFAR10, and CIFAR100. We compare the value rankings of our method with the existing state-of-the-art data valuation approaches data Shapley~\cite{ghorbani2019data}, LOO error, and Distributional Shapley~\cite{ghorbani2020distributional} and notice similar or better performance with significant speed-up in data valuation process.


\section{Related Work}
\textbf{Literature Review of Shapley Value.} Shapley value as formalized in~\cite{shapley1953value} establishes the axiomatic properties and demonstrates its unique ability to fairly allocate gains from cooperation among players. This seminal contribution laid the theoretical groundwork for subsequent developments in cooperative game theory~\cite{kalai1987weighted,grabisch1999axiomatic,myerson1977graphs,aumann2015values}. Shapley value has been extensively used for applications in economics,~\cite{gul1989bargaining,moulin1992application,roth1979shapley}, management science~\cite{kemahliouglu2011centralizing,dubey1981value}, online advertising~\cite{singal2019shapley}. In machine learning, it has been utilized for addressing the challenges in pricing ML training data, feature selection, and ML explainability.~\cite{cohen2005feature,zaeri2018feature} proposed to employ Shapley value properties for feature selection.~\cite{agarwal2019marketplace,fernandez13data} use Shapley value in market mechanism to price training data and match buyers to sellers data marketplace design.~\cite{lundberg2017unified} introduced the SHAP framework, leveraging Shapley values to provide interpretable explanations for machine learning models. Other works have also explored its utility in explaining black-box model predictions~\cite{chen2018shapley,sundararajan2020many,ghorbani2020neuron,bax2019computing}.\par

\textbf{Data Valuation in ML.} Recently, the subfield of data valuation in ML models has attracted significant attention and the existing works have shown promising outcomes. Data Shapley~\cite{ghorbani2019data,jia2019towards} proposed to use Shapley value from cooperative game theory for valuation of training data. KNN Shapley~\cite{jia12efficient} improved the efficiency of data Shapley by using a k-nearest neighborhood model. Distributional Shapley~\cite{ghorbani2020distributional} expanded the scope of valuation to the underlying data distribution instead of only considering the data points. Beta Shapley~\cite{kwon2022beta} relaxes the efficiency axiom in DataShapley and reports utility of data valuation in detecting mislabeled images in the training data. Data Banzhaf~\cite{pmlr-v206-wang23e} propose to estimate the Banzhaf value to improve results on noisy label detection. Several works have attempted to improve the efficiency of the Shapley value computation through approximation techniques~\cite{,kwon2021efficient}. Apart from this, other aspects of data value has been studied in~\cite{ilyas2022datamodels,nohyun2022data,covert2021explaining,covert2021improving,wang2020shapley}. However, approximation of Shapley value still remains a computationally expensive process, making it difficult to adapt for large models and datasets. The main goal of this work is to develop an alternative efficient data valuation framework to overcome this problem.\par

\textbf{Literature Review of Production Functions.}~\cite{mishra2007brief} offers a detailed outline of the evolution and econometrics of the production function. Aggregate production functions are used in macroeconomics to represent the relationship between total output of an economy (GDP) and the inputs used to produce that output. These inputs typically include capital ($K$), labor ($L$), and sometimes other factors like technology or natural resources \cite{barro1997macroeconomics, shephard2015theory}. The simplest production function used in economics, is the Cobb-Douglas production function introduced by~\cite{arrow1961capital}.~\cite{khatskevich2018production} identifies all multi-factor production functions with given elasticity of output and
from given elasticity of production. Production functions have been used in various domains, including health, education, and energy, to name a few~\cite{hanushek2020education,allan2009economics,bloom2004effect}. In our study, we adopt the concept of a production function and adapt it for data valuation. This approach draws inspiration from foundational works and recent advancements in the field.~\cite{jones2020nonrivalry} develops a theoretical framework that applies the production function to the economics of data, particularly employing data as an input for training machine learning models. Moreover,~\cite{farboodi2021model} highlights the role of data as information aimed at reducing forecast errors, which hints at a production function characterized by bounded returns to data. In our paper, we align with these perspectives and further the discourse by specifically focusing on the application of the production function concept in the valuation of data.

\section{Preliminaries}

Let an ML model $M$, intended for a task $T$, is trained on a dataset $B$ of size $m$. Let $U$ denote the performance metric and $U^T$ denote the performance obtained on task $T$. The overall performance $U$ is achieved after training a sufficient number of epochs $e$. Here the sufficient number of epochs means $|U_{e+i+1} - U_{e+i}| < \gamma$ for all $i \geq 0$, where $\gamma$ is an arbitrarily small value. It should be noted that $\gamma$ arises due to the randomness within the learning algorithm and not further training. The value of a data point is denoted by $\Phi$.\par

\textbf{Leave-One-Out (LOO) Error.} The LOO error computes the value of a datum $z$ based on the increase in performance obtained by adding it to the training set:
\begin{equation}
\Phi_{LOO}(z; U, B) = U(B) - U(B\setminus\{z\}).
\end{equation}

It struggles in differentiating data quality when similar data samples exist in the dataset. For example, if each sample has a duplicate copy in the dataset, the LOO will return a value $0$ for all of the samples. Shapley value overcomes this limitation by checking the marginal distribution over many subsets of the dataset.



\textbf{Shapley Value.} Shapley value~\cite{ghorbani2019data} measures the value of a data point $z$ as the weighted average of the performance increase when $z$ is added to different subsets of the dataset $B$:
\begin{equation}
\Phi_{s}(z;U, B) = \frac{1}{m}\sum_{k=1}^{m}{\frac{1}{\binom{m-1}{k-1}}}\sum_{S\subseteq B\setminus\{z\}} {\Delta(z; U, S)},
\label{eq:shapley_base}
\end{equation}
where $|S| = k-1$ for $k\in N$ and $\Delta(z; U, S) = U(S\cup \{z\}) - U(S)$. Thus, data Shapley value is the weighted average of the marginal contribution $\Delta(z; U, S)$. It satisfies the following Shapley value axioms:
\begin{itemize}
    \item \textbf{Dummy Player}: If $U(S \cup \{z\})=U(S) + e$ for all $S \subseteq B\i$ and some $e \in R$, then $\Phi(z; U, B) = e$.
    \item \textbf{Symmetry}: If $U(S \cup \{z\}) = U(S \cup \{z'\})$ for all $S \subseteq B\backslash\{z, z'\}$, then $\Phi(z; U, B) = \Phi(z'; U, B)$. 
    \item \textbf{Linearity}: $\Phi(z;\alpha_{1}U_{1}+\alpha_{2}U_{2},B)=\alpha_{1}\Phi(z;U_{1},B)+\alpha_{2}\Phi(z; U_{2}, B)$ for $\alpha_1,\alpha_2 \in R$. 
    \item \textbf{Efficiency}: $\sum_{z \in N}\Phi(z; U, B) = \Phi(U, B)$.
\end{itemize}
Further details regarding the interpretation of the above axioms in the context of machine learning can be referred to~\cite{ghorbani2019data} and~\cite{jia2019towards}.

\textbf{Production Function.} In economics, a production function expresses the relationship between the specific quantities and combinations of different inputs a company uses and the amount of output it produces. Commonly used production functions include Linear, Leontief, Cobb–Douglas~\cite{cobb1928theory,blume2008new}, CES, and CRESH~\cite{sickles2019measurement}, each varying in their assumptions for the input and the output. The widespread usage of the Cobb-Douglas production function is attributed to its simplicity and adaptability. It assumes homogeneity of inputs and this principle is consistent with many machine learning setups.

Let $P(g)$ denote the production over a set of goods $g = (g_1, g_2, .... g_n)$, the Cobb-Douglas production function is defined as
\begin{equation}
P(g) = A\prod_{i=1}^{n}g_{i}^{x_i},
\end{equation}
where $x_i$ is an elastic parameter for good $i$, and $A$ is the total factor productivity or the quality factor. If inputs are just labor $L$ and capital $K$, the production function is then
\begin{equation}
P = AL^xK^y. 
\end{equation}
It should be noted that the Cobb-Douglas production function also supports the diminishing returns in terms of both labor and capital. The~\textit{Law of Diminishing Returns}~\cite{shephard1974law} states that as the amount of a single factor of production is incrementally increased, the marginal output of a production process decreases. This property is analogous to how more data points have diminishing effects on a machine learning models performance. We therefore adapt the formulation of production functions in our proposed method to efficiently distribute the value of a cluster among its data members.\par


\section{Proposed Method}


A two-stage approach is proposed for efficient data valuation. First, data points are clustered together based on shared characteristics. Then, a leave cluster out (LCO) technique is applied to estimate the value of each cluster. This cluster value is then distributed among its members to obtain the preliminary individual data valuations. In the following, we delve into the building blocks of the proposed method and discuss its properties compared to the original Shapley methods.

\subsection{Leave-Cluster-Out}
\label{sec: LCO}
Cluster analysis is firstly performed on the given data and the marginal contribution of a cluster $c$ can be expressed as
\begin{equation}
V_c = U(B) - U(B\setminus c).
\end{equation}
The simple LOO error may provide an underestimated view of the true impact of specific data points, especially when similar data points remain in the dataset even after removal. Data Shapley alleviates this issue but suffers from high computational cost. By organizing data points into clusters based on their similarity, we ensure that when an entire cluster is removed, there are no closely-related points to mask the effect of its absence in other clusters. Consequently, this leads to a more precise assessment of the cluster's marginal contribution, effectively approximating its value. Furthermore, this clustering approach significantly reduces the number of model training iterations needed in comparison to Data Shapley since evaluations are conducted at the cluster level instead of for each individual data point. Once we have obtained cluster-level valuations, the subsequent step involves efficiently approximating the values of individual data points within each cluster.

\subsection{Value Propagation within a Cluster}
\label{sec:value_inside_cluster}
\textbf{Production Function for ML.} We adapt the Cobb-Douglas production function to approximate the data value for ML. In this context, we can draw an analogy: the labor $L$ corresponds to the available data points for the model; the learning capacity or the number of parameters in the model represents the capital $K$; and the final output is the obtained performance on the test set $U^T$. As both data quantity and model complexity exhibit diminishing returns, the Cobb-Douglas production function can be leveraged to effectively model learning performance. Therefore, we propose to approximate the model's performance after $e$ epochs as
\begin{equation}
U^T(S,N) = Af(S)h^T(N),
\label{eq:U-orig} 
\end{equation}
where $f(S)$ quantifies the informational utility of the dataset $S$ to the predictive efficacy of the model $U$, $T$ denotes the task, and $h^T(N)$ represents the effect of the model capacity which is dependent on $N$, the number of parameters of the model. 

Then, for a new point $z$, the performance change $\Delta U$ in the model incurred by the small increase ($\Delta S = \{z\}$) in $S$ can be computed by
\begin{align}
     & \Delta U^T(S,N) \notag\\
    =& A f(S+\Delta S)h^T(N) - A f(S)h^T(N)\notag\\
    =&A \left[f(S+\Delta S) -  f(S)\right] h^T(N)\notag\\
    =&A \left[\frac{f(S+\Delta S) -  f(S)}{o(z)}\right] h^T(N)o(z). \label{eq:diff-f}
\end{align}
To better understand Eq.~(\ref{eq:diff-f}), let us consider $f$ as a smooth function of $x$ as specified in Eq.~(\ref{eq:U-orig}), i.e., $U^T(x,N) = Af(x)h^T(N)$. Thus, a minor change in  $x$ leads to a change in $U^T$, which can be approximated by $A f'(x) h^T(N)\Delta x$. This allows us to interpret the expression enclosed in square brackets of Eq.~(\ref{eq:diff-f}) as effectively serving as the derivative of $f$ with respect to the set $S$, especially when considering incremental changes to $S$.\par

Also, in Eq.~(\ref{eq:diff-f}), $o(z)$ serves as an indicator of how a single data point enhances the model's overall performance and is a proxy to $\Delta x$ discussed above. Therefore, the difference $f(S + \Delta S) - f(S)$ captures the marginal impact on the model's performance when dataset $S$ is augmented by a new data point. Analogous to the concept of derivatives in calculus, this difference, when normalized by the contribution $o(z)$ of the individual point, can be interpreted as the \lq\lq rate-of-change\rq\rq of $f$ upon the addition of a new data point. This rate is contingent on both the existing dataset $S$ and the new data point being added. That is
\begin{equation}
U(S\cup \{z\}) - U(S) = \alpha^T(z)\beta(z, S),
\end{equation}
where 
\begin{align*}
    \alpha^T(z) = & Ah^T(N)o(z),\\
    \beta(z, S) = & \frac{f(S+\Delta S) -  f(S)}{o(z)}.
\end{align*}
Substituting the above into Eq.~(\ref{eq:shapley_base}) then gives
\begin{align} 
\Phi_{s}(z;U^T, B) 
  &= \frac{1}{m}\sum_{k=1}^{m}{\frac{1}{\binom{m-1}{k-1}}}\sum_{\substack{S \subset B \setminus \{z\}\\|S| = k-1}}{\alpha^{T}(z)\beta(z, S)} \notag\\
  &= \alpha^{T}(z) \beta^{*}(z, B)
\end{align}
where
\begin{align}
\beta^*(z, B) = \frac{1}{m}\sum_{k=1}^{m}{\frac{1}{\binom{m-1}{k-1}}}\sum_{\substack{S \subset B \setminus \{z\}\\|S| = k-1}}{\beta(z, S)}
\end{align}


\begin{prop}
\textbf{(Production Function Based Valuation for ML)}. Let $\alpha^T(z)$ denote the \textit{intrinsic value} of a datum $z$, i.e., $\alpha^T(z)$ is only dependent on the characteristics of $z$. The interaction of $z$ with rest of the data points in $B$ is captured by $\beta^*(z, B)$. From equitable properties of data valuation in~\cite{ghorbani2019data}, we postulate that for every datum $z$ having an intrinsic value $\alpha^T(z)$, the $\beta^*(z, B)$ acts as a multiplier or \textit{extrinsic factor} that decreases the value of $z$ if similar data points are present in the dataset. Similarly, it increases the data value if $z$ is a unique datum. Then the data valuation can be performed as below
\begin{equation}
\label{eq:Shapley_alpha_beta} 
\Phi(z;U^T, B) = \alpha^T(z)\beta^*(z, B).
\end{equation}
To simplify notation, we denote $\alpha^T(z)$ with $\alpha(z)$, and $\Phi(z; U^T, B)$ with $\Phi(z; U, B)$ for the rest of the discussion, since $T$ is invariant.
\end{prop}

\textbf{Fast Data Valuation.} Based on the above setup, we propose an efficient data valuation method that also works as an efficient proxy to Distributional Shapley~\cite{ghorbani2020distributional} to predict valuation for unseen data-points in the distribution. The existing Data Shapley adheres to two fundamental axioms~\cite{wang2023data}: symmetry and efficiency. \textit{Symmetry} states that for points $z$ and $z'$ that contribute similarly to the model's performance should have the same value, i.e. $U(S\cup\{z\}) = U(S\cup\{z'\})$ for all $S \in B\setminus\{z, z'\}$. \textit{Efficiency}, on the other hand, ensures that the aggregate value of all data points aligns with the overall performance achieved after training on the entire dataset. 

\begin{prop}
\textbf{(Fast Data Valuation of Cluster Data Members)} The \textit{symmetry} and \textit{efficiency} properties when applied to a specific cluster implies the data points within a cluster, characterized by similar features, will likely possess similar values and a cluster's value can be accurately represented as the sum of its constituent data points' valuations. 

Let $V_c$ ($=\Phi_c$) be the value of cluster $c$, the initial value assigned to any data point $z_i$ within this cluster is:
\begin{equation}
\label{eq: avg}
V_i = V_c/n_c,
\end{equation}
where $n_c$ is the number of data points in cluster $c$. Using this cluster-level assignment of initial data value, we estimate the actual data value based on Eq.~(\ref{eq:Shapley_alpha_beta}) as
\begin{equation}
\label{eq:vi_eq}
V_i^* = \alpha_i\beta_{i}^*.
\end{equation}
\end{prop}

\textbf{Estimating $\alpha$ and $\beta^*$.} Assuming each cluster contains an equal number of data points, the distribution of similar and dissimilar samples encountered by each datum becomes roughly uniform. This results in a near-constant \textit{extrinsic factor}, $\beta^*(z, B)$, across all data points. Thus, the value of these data points are directly proportional to $\alpha(z_i)$. We use $Q_i$ to denote the value of individual datum to differentiate it from $V_i$ value that is initialized by the cluster value in Eq.~(\ref{eq: avg}).


\begin{theorem}
For data point $z_i$, assuming there is no error in $\beta_{i}^*$, its adjusted value $V_i^{\Delta\alpha_i}$ is 
\begin{equation}
V_i^{\Delta\alpha_i} = (\alpha_i + \Delta\alpha_i)\beta_{i}^*
=\Gamma_{\alpha_i}\alpha_i\beta_{i}^*,
\label{eq:corrective_alpha}
\end{equation}
where $\Gamma_{\alpha_i}$ is an adjustment factor for $\alpha_i$ 
\begin{equation}
\Gamma_{\alpha_i} = 1 + \frac{Q_i}{\sum_{z_j\in c}Q_j}V_c.
\label{eq:adjustment_factor}
\end{equation}
\end{theorem}

The adjustment factor $\Gamma_{\alpha_i}$ ensures that value of individual data point is adjusted not only based on its personal contribution (\(Q_i\)) but also in proportion to the corresponding cluster's overall impact (\(V_c\)). This dual consideration is crucial for accurately reflecting the true value of each data point, balancing internal cluster contributions with the broader context of the cluster’s role within the entire dataset. Including \(V_c\) in the adjustment ensures that the recalibration of \(\alpha_i\) remains sensitive to both intra-cluster dynamics and the comparative significance of each cluster, providing a more nuanced and equitable valuation of data points across a heterogeneous dataset.

\begin{corollary}
When all data points in a cluster are exactly the same, the adjustment factor should be equal to $1$ so that for each point in $c$, the value becomes $V_i$. But the above formulation of $\Gamma_{\alpha_i}$ yields $1 + 1/n$ when all the points are identical as $V_i$ and $V_j$ will be equal for any $i$, $j$. Thus, we normalize $\Gamma_{\alpha_i}$ as follows
\begin{equation}
\Gamma_{\alpha_i} = \frac{1}{1 + V_c/n_c} \left(1 + \frac{Q_i}{\sum_{z_j\in c}Q_j}V_c \right).
\end{equation}
\end{corollary}

Similar to $\alpha_i$, we find the adjustment factor for $\beta_{i}^*$, i.e. $\Gamma_{\beta_{i}^*}$. $\beta_{i}^*$ measures the interaction of $z_i$ with all other data points in $B$. As all data points similar to $z_i$ belong to the same cluster and $\beta_{i}^*$ is only affected by the other members in $z_i$'s cluster. We use the distance between $z_i$ and cluster centroid as a measure to it's belongingness to the cluster or similarity to other points in the cluster. 

\begin{theorem}
For data point $z_i$, assuming no error in $\alpha_i$, its adjusted value $V_i^{\Delta\beta_i^*}$ is
\begin{align}
V_i^{\Delta\beta_{i}^*} = \alpha_i(\beta_{i}^* + \Delta\beta_{i}^*) = \Gamma_{\beta_{i}^*} \alpha_i\beta_{1i},
\label{eq:corrective_beta}
\end{align}
where $d_i$ is the distance of $z_i$ and $\Gamma_{\beta_{i}^*}$ is the adjustment factor represented as
\begin{equation}
\Gamma_{\beta_{i}^*} = \frac{1}{1 + V_c/n_c} \left(1 + \frac{d_i}{\sum_{z_j\in c}d_j}V_c \right).
\end{equation}
\end{theorem}

\textbf{Production Function based Data Value Estimation.} The final approximation value $\hat{\Phi}_i$ of the data point is 
\begin{align}
    \hat{\Phi}_i = &(\alpha_i + \Delta\alpha_i)(\beta_{i}^* + \Delta\beta_{i}^*).
\end{align}
Ignoring $\Delta\alpha_i\Delta\beta_{i}^*$ then gives
\begin{align}
        \hat{\Phi}_i\approx &(\alpha_i + \Delta\alpha_i)\beta_{i}^* + \alpha(\beta_{i}^* + \Delta\beta_{i}^*) - \alpha_i\beta_{i}^*.
\end{align}
By substituting Eq.~(\ref{eq:vi_eq}), Eq.~(\ref{eq:corrective_alpha}), Eq.~(\ref{eq:corrective_beta}), we obtain.
\begin{align}
\hat{\Phi}_i = & V_i^{\Delta\alpha_i} + V_i^{\Delta\beta_{i}^*} - V_i\notag\\
             = & V_i(\Gamma_{\alpha_i} + \Gamma_{\beta_{i}^*} - 1) \notag\\
             = & V_i \left[ \left(\frac{1}{1 + V_c/n_c} \right) \left(1 + \frac{Q_i}{\sum_{z_j\in c}Q_j}V_c \right) + \right.\notag\\
               &\left. \left(\frac{1}{1 + V_c/n_c} \right) \left(1 + \frac{d_i}{\sum_{z_j\in c}d_j}V_c \right) - 1 \right].
\label{eq:corrected_shap}
\end{align}
For the reader's convenience, Algorithm~\ref{alg:fast_data_shapley} outlines the implementation steps of the EcoVal efficient data valuation framework.

\subsection{Discussion: Comparison with Original Shapley}
Let $E(z)$ denote the appropriate embedding from a machine learning model or the pre-final layer of a deep learning model for a data point $z$. We extend the notion of \textit{Lipschitz Stability} of data Shapley introduced in \cite{ghorbani2020distributional} to estimate the difference in value of different data points. We use proximity of the embeddings $E(z)$ as a proxy to the closeness in the underlying data distribution and formalize the same in the following Theorem.

\begin{theorem}
For any $z_j$, $z_k$ if $||E(z_j) - E(z_k)|| < \epsilon$ then, $|\Phi(z_j) - \Phi(z_k)| \leq \epsilon_1$ for very small $\epsilon, \epsilon_1 \geq 0$ 
\end{theorem} 

From the principle of clustering, a datum $z_j$ belongs to cluster $c$ if 
\begin{align}
||E(z_j) - E(z_k)|| \leq \epsilon, \forall z_k \in c, \\
\intertext{then for this cluster}
|\Phi(z_j) - \Phi(z_k)| \leq \epsilon_1, \forall z_k,z_j \in c.
\end{align}

\begin{algorithm}[t]
\caption{EcoVal Data Valuation}
\label{alg:fast_data_shapley}
\begin{algorithmic}[1]
\State $M(.;\psi)$: Fully Trained Model
\State $B$: Training Dataset
\State $B_{D}$: Set of available points from the underlying distribution of $B$
\State $M_{-n}(x;\psi) \gets$ Embedding of data $x$ obtained from the $n^{th}$ last layer of the model
\State Let $E(x) = M_{-n}(x;\psi)$
\State Let $A_{c}$ be a clustering algorithm then $(x_i, c_j) \gets A_{c}(B_D) \forall x_i \in B_D$ where $c_j \in C$ is the cluster associated with $x_i$ and $C$ is the set of all clusters
\State Find valuation at cluster level \\ 
$V_{c_{j}} = U(B) - U(B\setminus c_j)$ $\forall c_j \in C$
\State Initialize value $V_i$ for each cluster member $x_i$ \\
$V_i = V_{c_{j}}/n_{c_{j}}$, where $n_{c_{j}}$ is the number of elements in cluster $c_j$ to which $x_i$ belongs
\State Initialize: $D \gets$ []
\For{$c_{j} \in C$} 
    \State Sample $X_j$ = \{$x_1^j$, $x_2^j$, ... $x_{n_{c}}^{j}$\} from $c_{j}$
    \State $D \gets D\cup X_j$
\EndFor

\State Run TMC Shapley~\cite{ghorbani2019data} \\
$(x_k, v_{TMC_{k}}) \gets TMC(U^T, D) \forall x_k\in D$
\State Train a regression model $R$ on the sampled data \{$(x_1, v_{TMC_{1}}), (x_2, v_{TMC_{2}})....(x_{\lvert D \rvert}, v_{TMC_{\lvert D \rvert}}) $\}

\For{$c_{j} \in C$} 
    \State $(x_i^j, q_i^j) \gets R(x_i^j) \forall x_i^j \in c_{j}$
    \State Let $\Bar{x}_{c_{j}}$ be the centroid of the cluster $c_j$ \\
    $(x_i^j, d_i^j) \gets distance(x_i^j, \Bar{x}_{c_{j}}) \forall x_i \in c_j$
\EndFor

\For{$x_i \in B$}
    \State Find correction term for $\alpha$ \\ $\Gamma_{\alpha_i} = \frac{1}{1 + V_{c_{j}}/n_{c_j}}(1 + \frac{q_i^j}{\sum_{z_k\in c_j}q_k^j}V_{c_{j}})$
    \State Find correction term for $\beta_{i}^*$ \\ $\Gamma_{\beta_{i}^*} = \frac{1}{1 + V_{c_{j}}/n_{c_j}}(1 + \frac{d_i^j}{\sum_{z_k\in c_j}d_k^{j}}V_{c_{j}})$
    \State Final valuation = $V_i*(\Gamma_{\alpha_i} + \Gamma_{\beta_{i}^*} - 1)$
\EndFor
\end{algorithmic}
\end{algorithm}


It means all Shapley values lie within an $\epsilon_1$ interval. Therefore, $\Phi(z_j)$ for any $z_j\in c$ can be expressed as 
\begin{equation}
\label{eq:value_in_cluster}
    \Phi(z_j) = \bar{\Phi}_c + \delta(z_j),
\end{equation}
where $\delta(z_j) \leq \epsilon/2$ and $\bar{\Phi}_c$ lies somewhere in the $\epsilon_1$ interval. 

\begin{corollary}
\label{thm:V_cluster_value}
The value $V_c$ of the a cluster by Shapley axioms is defined as
\begin{align}
V_c & = \sum_{z_j \in c} \Phi(z_j) = n_c\bar{\Phi}_c + \sum \delta(z_j).
\label{eq:cluster_value}
\end{align}
\end{corollary}
The detailed proof to \ref{thm:V_cluster_value} is provided in the Appendix.

\begin{theorem}
\label{thm:shapley_comparsion}
The difference between the original Shapley value and our proposed approximated data value is
\begin{align}
\Delta\Phi_i & \approx \frac{n_c\bar{\Phi}_c\delta_R}{\sum_{z_j\in c}Q_j} + \frac{n_c^2\bar{\Phi}_cQ_i\delta_R}{(\sum_{z_j\in c}Q_{j})^2}.
\end{align}
Due to the intrinsic limitations on the magnitudes of average Shapley value within a cluster $\bar{\Phi}_c$ and individual point contribution $Q_i$, both values inherently remain within a bounded range. As cluster size $n_c$ increases, the predicted aggregate value $\sum_{z_j\in c}Q_j$ proportionately grows, naturally restricting the potential expansion of $\frac{n_c}{\sum_{z_j\in c}Q_j}$. Additionally, a moderately accurate regression model ensures a low $\delta_R$ error. Therefore, our method produces Shapley value estimates $\Phi_i$ with minimal margin of error.

\end{theorem}

The detailed proof to \ref{thm:shapley_comparsion} is provided in the Appendix.

\section{Experiments}
We show the broad effectiveness of the proposed valuation framework and its general applicability to machine learning models through empirical evidence. We estimate the value of data in a machine learning model in MNIST, CIFAR10, and CIFAR100 datasets. We compare our method with Data Shapley~\cite{ghorbani2019data} and Distributional Shaply~\cite{ghorbani2020distributional}.\par
\textbf{Experiment Settings:} Following the common practice in previous works, we extract the features from last layer of a pre-trained network and apply Shapley on this embedded vector. We sample a small subset, i.e. 200 samples from the original training data and run the baseline methods TMC-Shapley (Data Shapley) and distributional Shapley. 2000 samples are used for testing and holdout for Shapley calculation. We keep 10,000 samples which are never seen by model or valuation method at any point, we call this \textit{out-of-sample (OOS) set}. The rest of the samples are used as data distribution and exposed to Distributional Shapley, and our method during the clustering step and $\alpha$ correction step. We use Gaussian Mixture Models (GMM) for clustering. Our proposed method works for both in-distribution and OOS samples. As Data Shapley only works for in-distribution samples, we compare our results with Distribution Shapley for out-of-sample data. We use Gaussian Mixture Model (GMM) clustering with default parameters of scikit-learn's implementation, covariance-type='full', tol=0.001, reg-covar=1e-06, max-iter=100, n-init=1, init-params='kmeans', and 30 mixtures/clusters.
\begin{figure}[t]
\centering
    \includegraphics[width=0.5\textwidth]{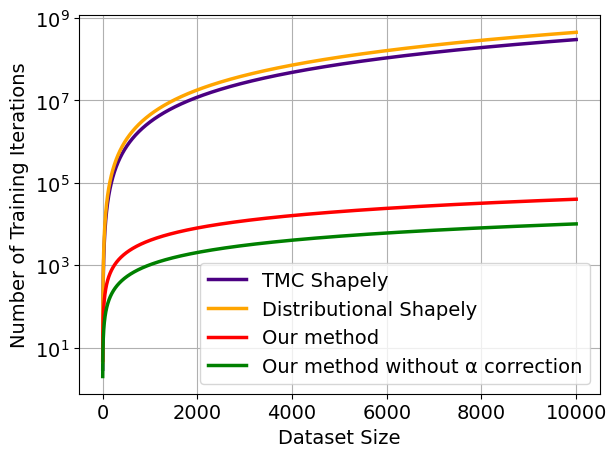}
\caption{Computation cost in terms of number of training iterations required for the given dataset size. We compare EcoVal with TMC Shapley (also known as Data Shapley), distributional Shapley, and a lighweight version of EcoVal. Our method requires substantially lower number of training iterations for data valuation.}
\label{fig:runtime-cmp}
\end{figure}
\subsection{Comparative Analysis of the Computational Time}
The Data Shapley approximation method TMC Shapley~\cite{ghorbani2019data} converges in approximately $3\lvert B \rvert$ (or $3\times m$ in Eq.~\ref{eq:shapley_base}) Monte Carlo samples. Each Monte Carlo sample is a random permutation of the data points in the training set. The marginal contribution of a data point $z$ in a given permutation is obtained as the performance difference between the model trained on data points before this datum, say $S$, and the model trained on $S\cup\{z\}$. Each point is added sequentially meaning $\lvert B \rvert$ training runs are required in a single Monte Carlo sample. This makes the number of training runs in the order of $O(\lvert B \rvert^2)$. Distributional Shapley's \cite{ghorbani2020distributional} time complexity is similar with $T$ runs to get an unbiased estimate using different subsets $S$ from the underlying data distribution. This makes the number of training runs of Distributional Shapley $O(T*\lvert B \rvert^2)$.\par

Our method performs clustering that takes less time than training a machine learning or deep learning model in most real-world scenarios. This is a one time effort, so the complexity is in the order of $O(1)$. Estimating the value of each cluster requires $O(p)$ training runs. Apart from that, our method involves running Data Shapley on a curated subset $p$ containing an equal number of points from each cluster, this take $O(p^2)$ time. The size of this subset $p$ is much smaller than $\lvert B \rvert$. The total number of training runs required is in the order of $O(1) + O(p^2) + O(p)$. We compare the computational cost associated with the TMC Shapley/Data Shapley with our EcoVal method. As illustrated in Figure~\ref{fig:runtime-cmp}, EcoVal requires significantly fewer training iterations, approximately \(10^3\) to \(10^5\), compared to \(10^7\) to \(10^9\) required by traditional TMC Shapley. Our method reduces computational overhead by employing a curated subset approach, which involves running Data Shapley on a subset \( p \) containing an equal number of points from each cluster. This curated subset approach requires computational resources in the order of \( O(p^2) \), significantly less than the \( O(|B|^2) \) required by the standard TMC Shapley method, where \( |B| \) represents the size of the full dataset. The size of subset \( p \) is much smaller than \( |B| \), which explains our method's efficiency and scalability. With the increase in the dataset size, the utility of our EcoVal becomes more evident. Our method without $\alpha$ correction is even faster with negligible loss in valuation quality.

\begin{figure}
\centering
    \includegraphics[width=0.34\textwidth]{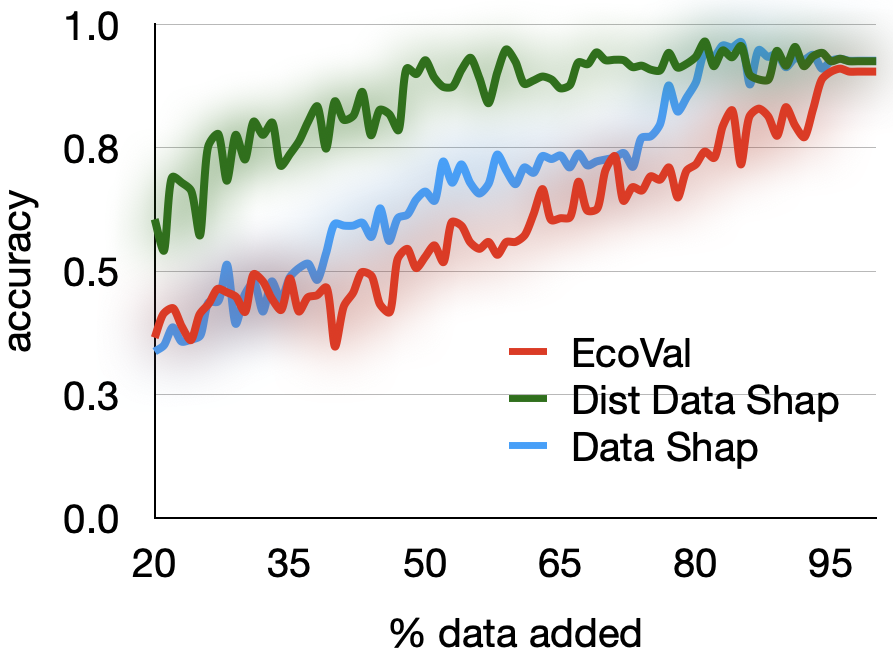}
    \includegraphics[width=0.34\textwidth]{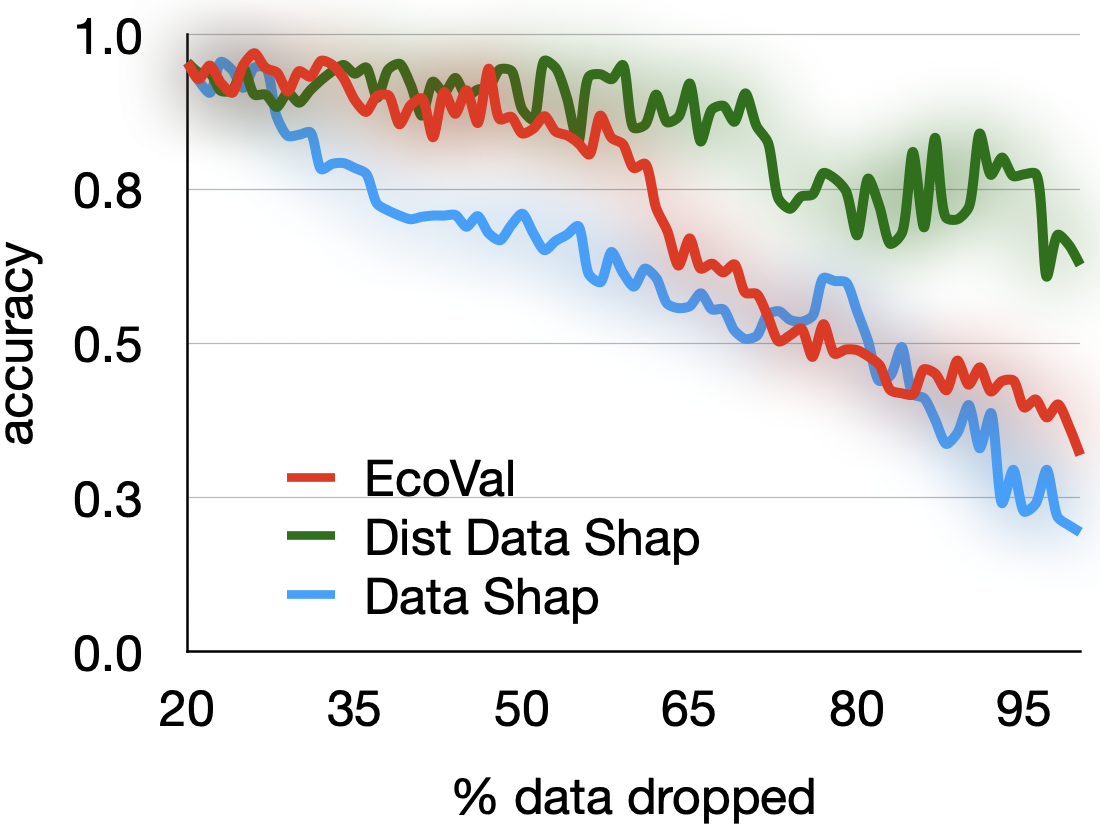}
    
    \includegraphics[width=0.34\textwidth]{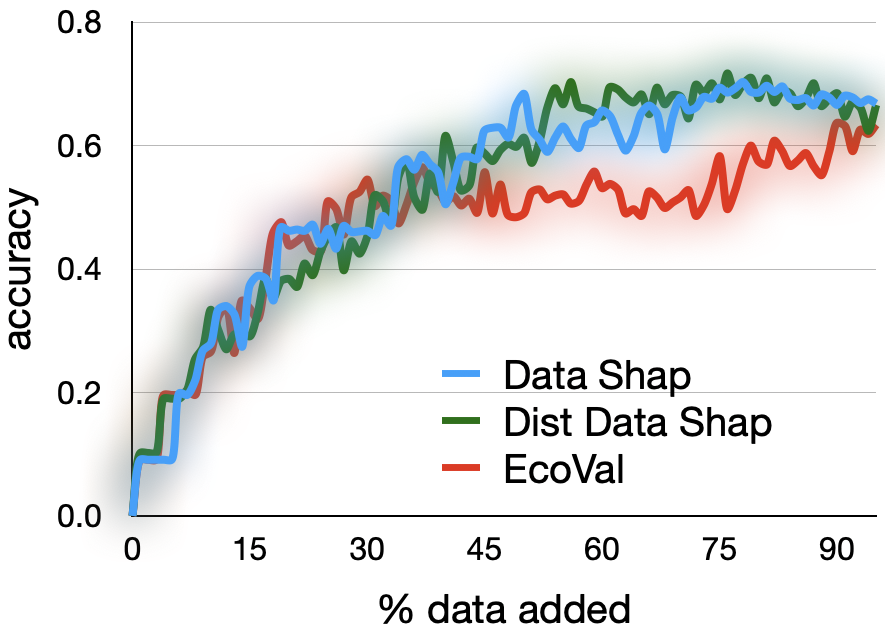}
    \includegraphics[width=0.34\textwidth]{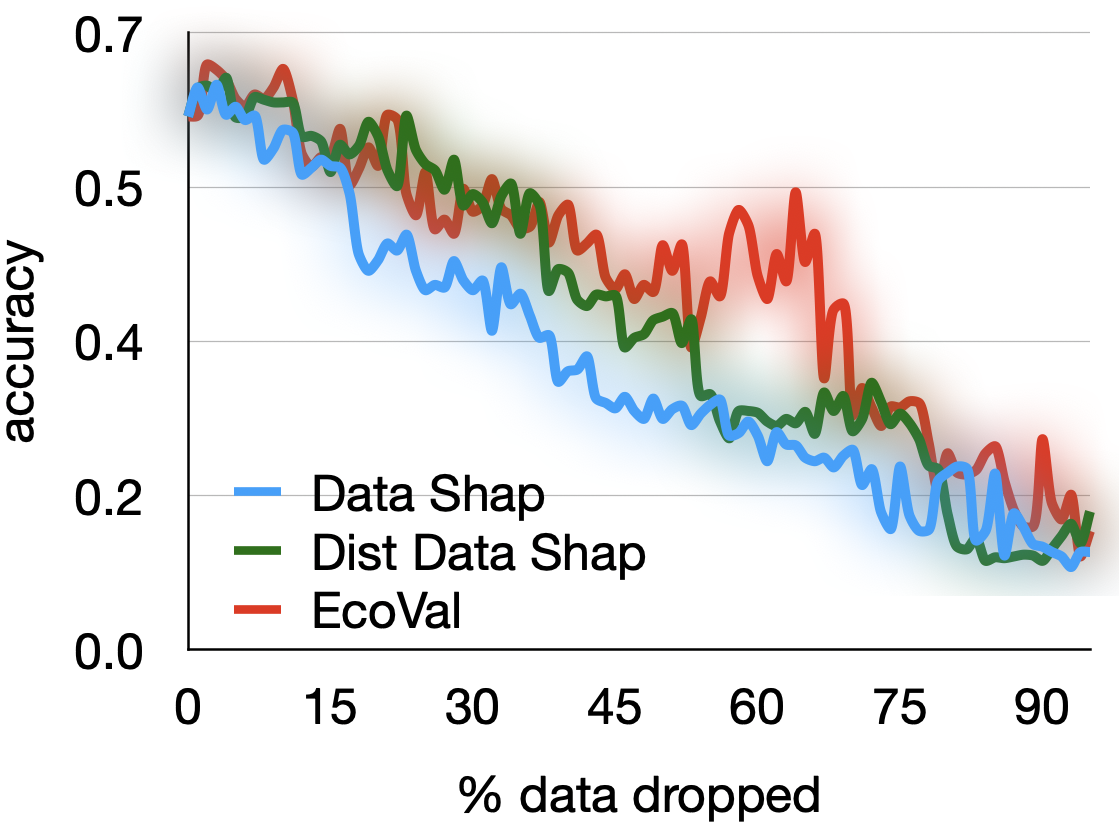}

    \includegraphics[width=0.25\textwidth]{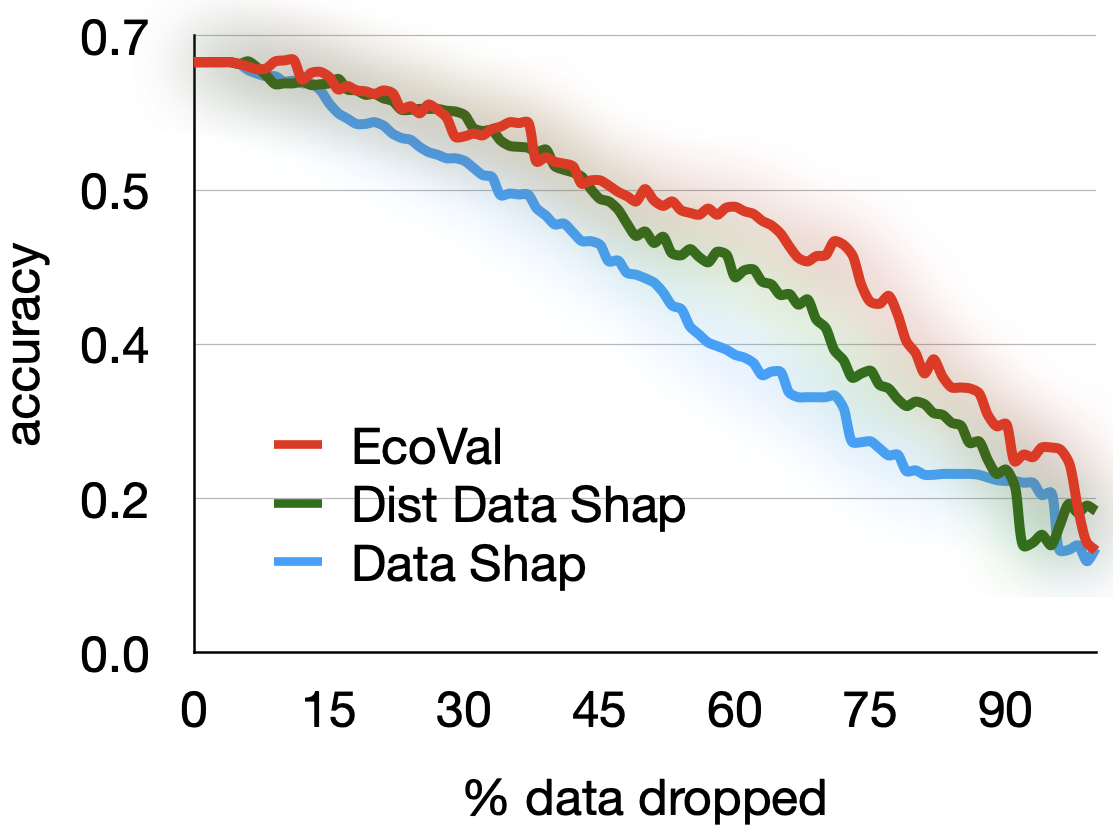}
\caption{Accuracy difference with respect to the \% of data points added or removed. We add or remove the highest valued data points first and then subsequently add or remove the lesser value data, respectively. The top, middle and bottom rows show the results for MNIST, CIFAR10, CIFAR100, respectively with in-distribution valuation. The EcoVal gives comparable or better performance when compared to Data Shapley and Distribution Data Shapley.}
\label{fig:cifar10-mnist-in-distribution}
\end{figure}

\begin{figure}[!h]
\centering
\includegraphics[width=0.3\textwidth]{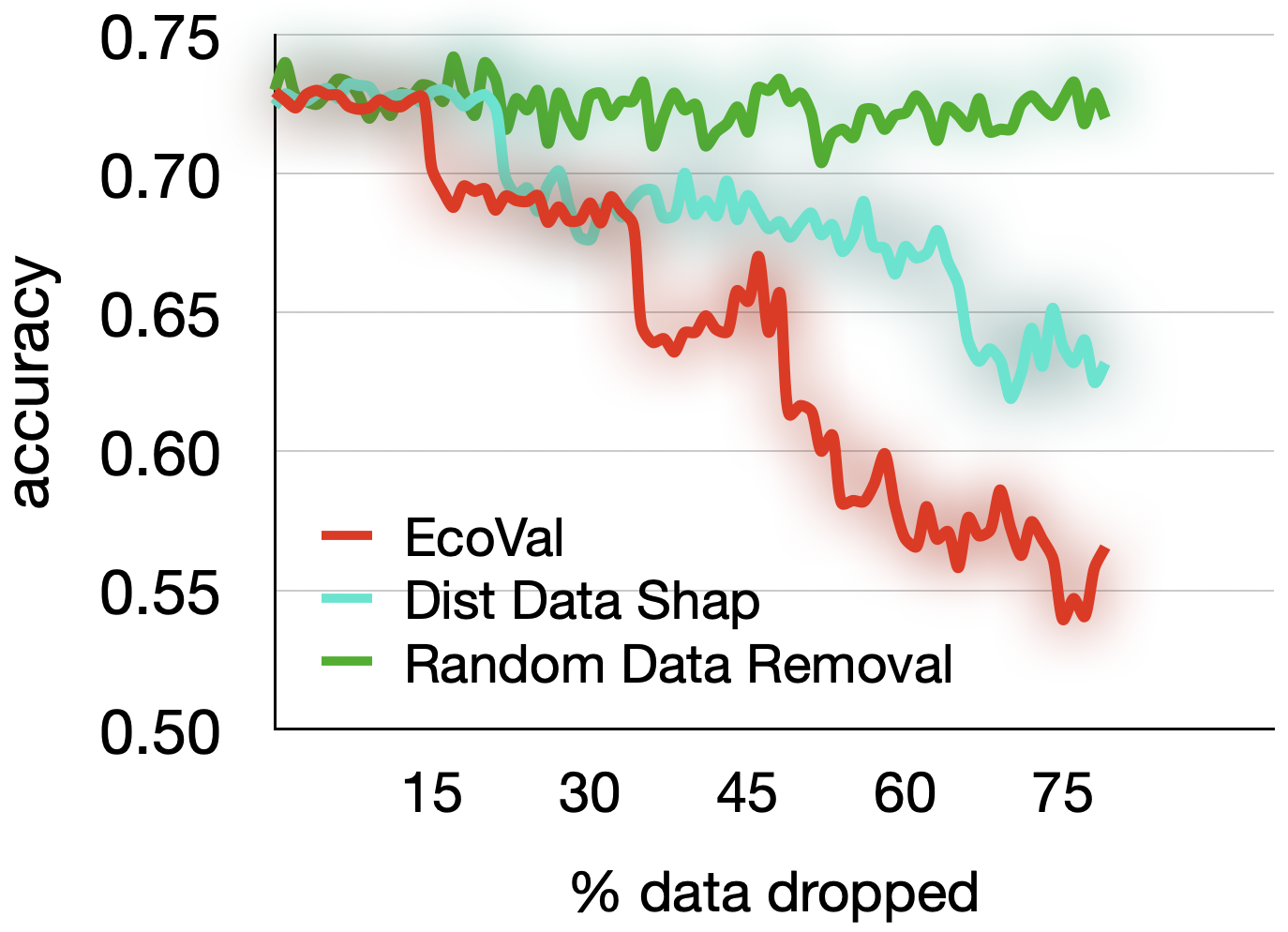}
\includegraphics[width=0.3\textwidth]{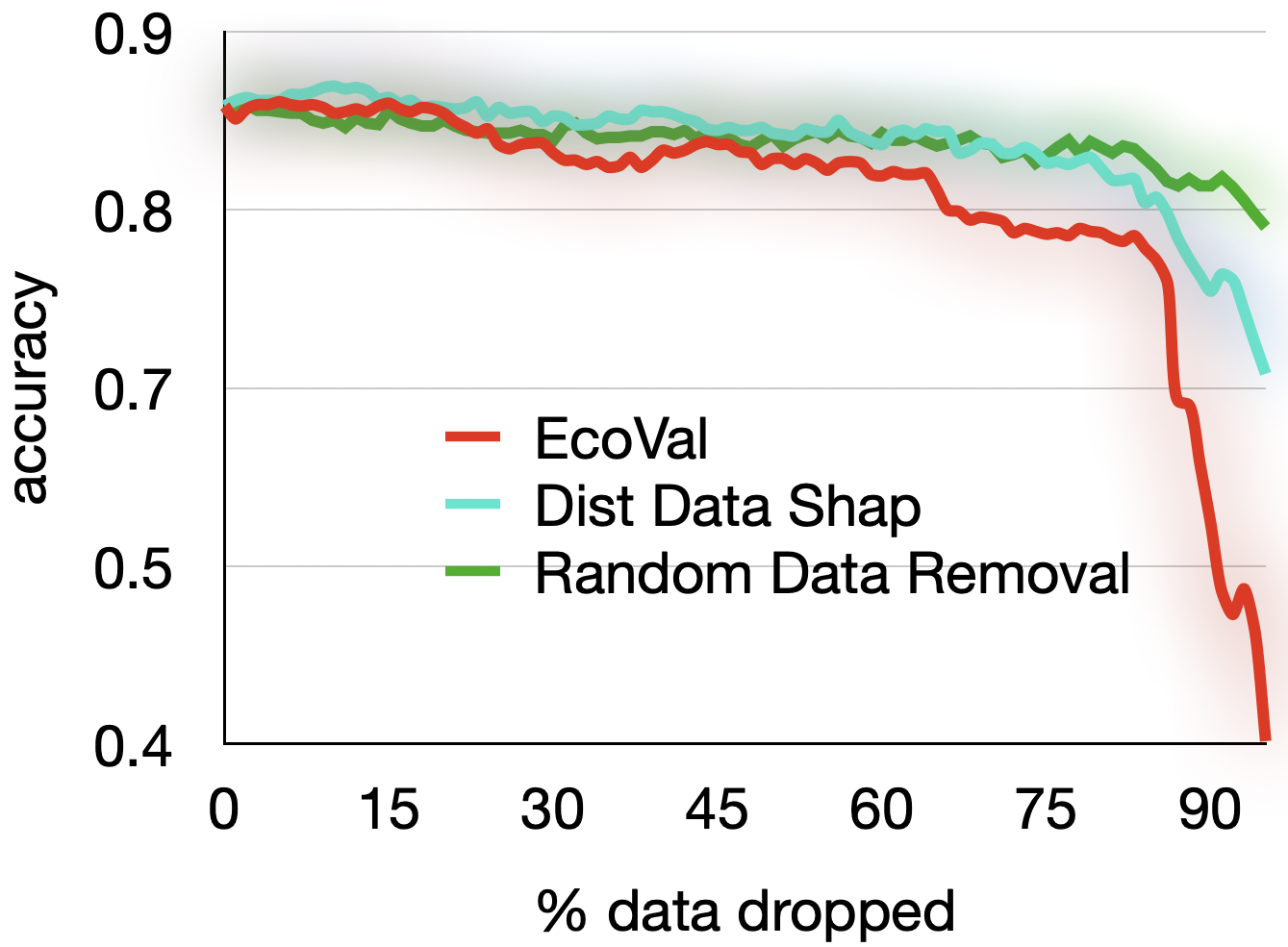}    \includegraphics[width=0.3\textwidth]{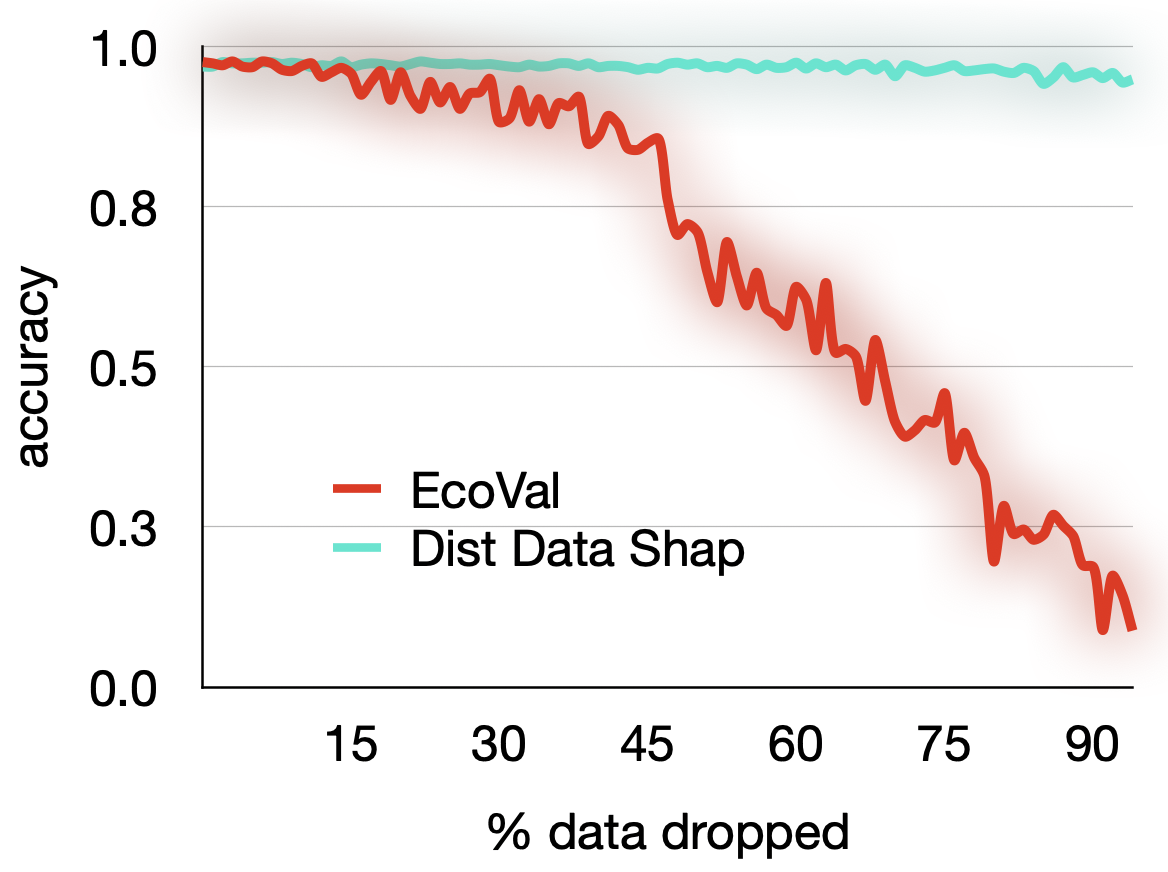}
\caption{Data valuation on out-of-sample data (top left to bottom right: CIFAR10, CIFAR100, MNIST, respectively). Our EcoVal method outperforms Distributed Data Shapley and Random Data Removal by getting steeper performance drop with increasing \% valuable data removal.}
\label{fig:cifar10-cifar100-mnist-oot}
\end{figure}
\subsection{Data Point Addition and Removal Experiments}
We evaluate the data valuation methods by running the data point addition and removal experiments as proposed in~\cite{ghorbani2019data}. For a given model and dataset, the data points are added in the order of predicted value, i.e. from largest to lowest values, and the model is retrained for each addition. Similarly, another experiment is conducted where we remove samples with high values and observe the performance drop. The impact of removal and addition of high value data-points help us measure the effectiveness of data valuation techniques. We compare our results with state-of-the-art Data Shapley and Distributed Data Shapley valuation methods.\par

\textbf{Removing most valued data points.} We predict values of data-points using each valuation method and we measure the drop in performance of model by removing most-valued data-points for each method. A better valuation method's high value data-points will result in a higher drop in performance. So, for removal of most valued points, the method resulting in higher performance drop is a better valuation method.\par

\textbf{Adding most valued data points.} This approach is vice-versa of the previous approach, we add most valued data-points into the training set and observe the increase in the performance. A higher increase on adding the top data-points shows better valuation method. Figure~\ref{fig:cifar10-mnist-in-distribution} shows the performance drop and increase upon adding and removing most valued points, respectively. It can be observed that EcoVal performance drop is slightly less than that of Data Shapley but significantly higher than the Distributed Data Shapley which is desired. Similar patterns can be observed in the data addition graph also.\par

\textbf{Removing most valuable data points from out-of-sample set.} 
The discussed earlier, EcoVal supports data valuation for out-of-sample data as well which is supported only by Distributed Data Shapley. Therefore, we compare the OOS valuation results between them. Figure~\ref{fig:cifar10-cifar100-mnist-oot} shows the performance drop by removing the most valuable points from an out-of-sample set of size $10,000$. It can be observed that EcoVal's performance drop is very high as compared to Distributed Data Shapley. The steep drop in the performance after removing the most valuable data points implies better precision for data valuation in our EcoVal framework.

\subsection{Effect of the Adjustment Terms}
\begin{figure}[t]
\centering
    \includegraphics[width=0.34\textwidth]{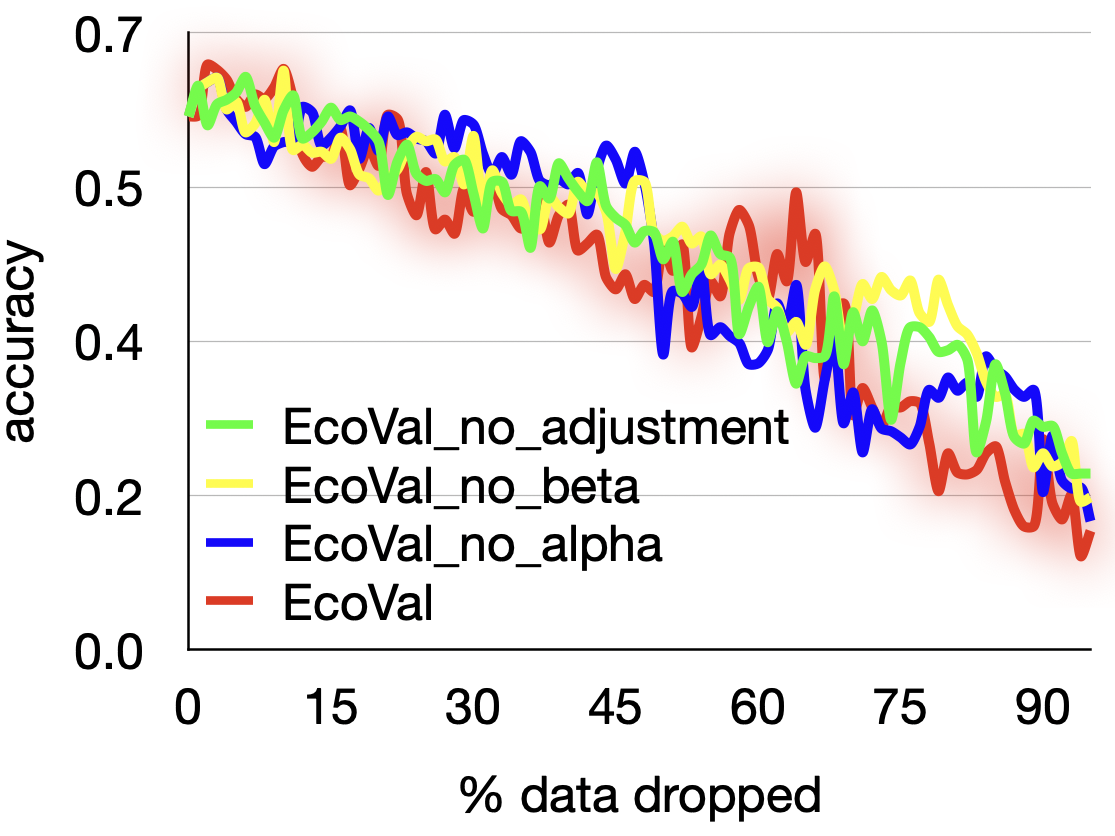}
    \includegraphics[width=0.34\textwidth]{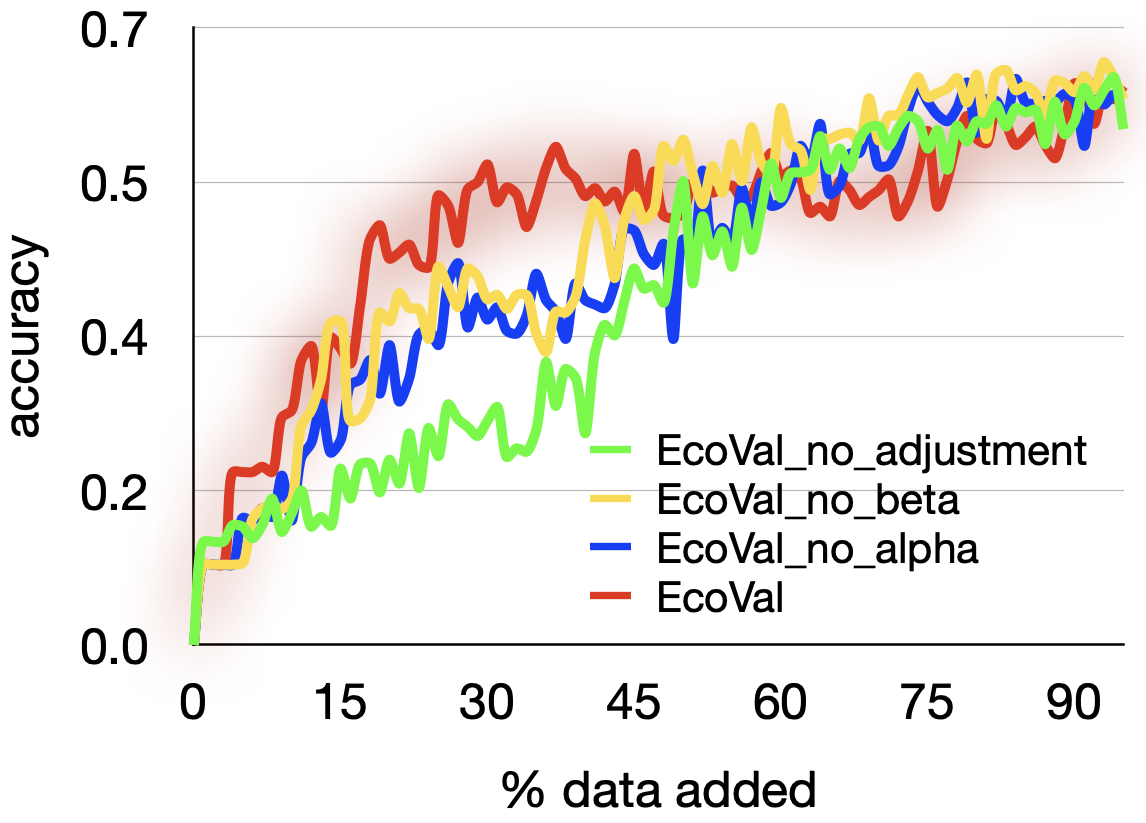}
\caption{Effect of adding different adjustment terms (refer Section~\ref{sec:value_inside_cluster}). EcoVal: the full proposed method, EcoVal\_no\_alpha: removal of $\alpha$ adjustment term, EcoVal\_no\_beta: removal of $\beta$ adjustment term, EcoVal\_no\_adjustment: EcoVal without any adjustment terms, i.e. the mean of the cluster value used as the data value.}
\label{fig:cifar10-ablation}
\end{figure}

We observe the effect of removing different adjustment terms $\Gamma_{\beta_{i}^*}$, $\Gamma_{\alpha_{i}}$ or both in the EcoVal framework and show the results in Figure~\ref{fig:cifar10-ablation}. The overall EcoVal framework with the terms $\alpha$ and $\beta$ performs the best, in general. Eliminating one of the adjustment terms deteriorates the quality of the valuation by a small margin. Removing both corrections significantly impacts the quality of data valuation. This is particularly visible in the initial phase of adding the most significant data points. It should be noted that eliminating $\Gamma_{\alpha_{i}}$ only affects the valuation quality marginally, but completely removes the need for model training, giving an even more efficient version of our valuation method.\par

The performance of different variations of our method (EcoVal, EcoVal-no-$\alpha$, EcoVal-no-$\beta$, EcoVal-no-adj) would vary depending on the intra-class and inter-class variations present in a dataset. In Figure 4, the performance differences are not sometimes consistent. We report our observation based on the experiments with CIFAR-10 dataset. More complex datasets with larger variations may better reveal the impact of these adjustment factors, which is the future scope of this work.

\subsection{Mutual Influence of Clustering Methods and Adjustment Terms}
The effectiveness of EcoVal is intrinsically related to the success of the clustering method, as the initial valuation is carried out at the cluster level. The adjustment terms are meant to marginally correct the data value distribution within each cluster. The similarity within a single cluster compared to the similarity between clusters significantly influences the structure and application of the adjustment factors. We assume that data points within the same cluster have a higher degree of similarity compared to points across different clusters, a standard assumption in clustering algorithms such as Gaussian mixture models (GMM). However, the variance in the degree of similarity across different clusters (inter-cluster dissimilarity) justifies the need for a scaling factor such as $\Gamma_{\alpha_i}$, introduced in Equation~\ref{eq:adjustment_factor}. This accounts for the relative contribution of an entire cluster to the model's performance, acknowledging that some clusters may be more pivotal due to their positioning, density, or the nature of the data points they contain. This process depends on the intra-cluster variation present in a dataset. For some dataset, a simple distribution of the value of the cluster without correction factors can also give a good valuation, leaving a minimal scope of correction.

\section{Conclusion}
This work presents a focused study on improving the speed of data valuation in machine learning models. We develop an efficient data valuation method that is significantly fast and practical for working with large datasets. Our method works for both in-distribution and out-of-sample data. The proposed EcoVal data valuation framework shows comparable and sometimes even better results than the existing approaches for in-distribution data. For out-of-sample data points, our method significantly outperforms competing methods, thus establishing a new state-of-the-art. This proves our method's utility in a data market, where new data points analogous to our out-of-sample set are generated every passing instant. Our valuation also shows negligible error margin with the vanilla Shapley value approximation. The points mentioned above collectively make the proposed method a robust and scalable approach to estimate the value of data across a variety of machine learning models.

\section*{Acknowledgment}
This research/project is supported by the National Research Foundation, Singapore under its Strategic Capability Research Centres Funding Initiative. Any opinions, findings and conclusions or recommendations expressed in this material are those of the author(s) and do not reflect the views of National Research Foundation, Singapore.\par
This research is also supported by the Department of Science and Technology-Science and Engineering Research Board (DST-SERB), India project under Grant SRG/2023/001686. 
{
\bibliographystyle{unsrt}
\bibliography{main}
}
\newpage 
\appendix
\setcounter{page}{1}
\section*{Appendix}

\section{Proofs}
\subsection{Theorem 3.4}
Let equal number of samples are used from each cluster to run TMC Shapley. Then the \textit{intrinsic value} of a datum is independent of proportion and bias in the data distribution. If $n_s$ such samples exist, the value is divided into these $n_s$ samples. From Shapely axioms, the data Shapley at the current stage of TMC becomes approximately $\frac{\alpha_i}{n_s}$.\par

For rest of the samples in the TMC, we train a regression model $R$ for predicting $\frac{\alpha_i}{n_s}$ for an input data. If the predicted Shapley for any $z_i \in c$ is $Q_i$, then assuming no error is introduced due to the TMC Shapely algorithm, this gives us
\begin{equation}
Q_i = \frac{\alpha_i}{n_s} + \delta_{R_i},
\end{equation}
where $\delta_{R_i}$ is error introduced by the regression model. $Q_i$ denotes $b*\alpha_i$, where $b$ is some constant. We use this to obtain the adjustment factor for $\alpha_i$ in Eq.~\ref{eq:corrective_alpha}. Assuming $V_i$ and $d_i$ do not have any error, we take the differentiation of Eq.~\ref{eq:corrected_shap} with respect to $z$
\begin{equation} 
\label{eq:eq_differentiate}
    \frac{\partial\Phi_i}{\partial z} = \frac{1}{\partial z} (\frac{V_i}{1 + V_c/n_c})\left[\frac{n_c \partial Q_i}{\sum_{z_j\in c}Q_j} - \frac{n_cQ_i \partial(\sum_{z_j\in c}Q_j)}{(\sum_{z_j\in c}Q_j)^2}\right].
\end{equation}
\textbf{Intuition.} $V_i$ does not have any error as this is the difference between the performance with and without the cluster $c$ divided by a constant. Both the values can be directly computed from the model. Similarly, $d_i$ is the distance of the datum from the centroid of the cluster $c$ which can be calculated without any error.

Comparing Eq.~\ref{eq:eq_differentiate} with the change in the Shapley value leads to the following inequality.

\begin{align} 
\label{eq:eq_differentiate_compare}
\Delta\Phi_i & \leq \frac{V_c/n_c}{1 + V_c/n_c}\left[\frac{\delta_Rn_c}{\sum_{z_j\in c}Q_j} + \frac{n_cQ_i(n_c\delta_R)}{(\sum_{z_j\in c}Q_j)^2}\right], \nonumber\\
\intertext{where $\delta_R = \max\limits_{i}\delta_{R_i}$ is the maximum error of the regression model}
\Delta\Phi_i & \leq V_c\left[\frac{\delta_R}{\sum_{z_j\in c}Q_j} + \frac{n_{c}Q_{i}\delta_R}{(\sum_{z_j\in c}Q_j)^2}\right],\nonumber\\
\intertext{as $V_c\geq0$, $n_c\geq1$ therefore, $\frac{1}{1+V_{c}/n_{c}} \leq 1$. 
From Eq.~\ref{eq:value_in_cluster} and Eq.~\ref{eq:cluster_value}}
\Delta\Phi_i & \leq (n_c\bar{\Phi}_{c} + n_c\epsilon/2)\left[\frac{\delta_R}{\sum_{z_{j}\in c}Q_j} + \frac{n_{c}Q_{i}\delta_R}{(\sum_{z_j\in c}Q_j)^2}\right].\nonumber\\
\intertext{Ignoring factors with multiples of $\epsilon$ and $\delta$ as these values are very small. We get the final difference between the original Shapley value and our proposed approximated Value as below}
\Delta\Phi_i & \approx \frac{n_c\bar{\Phi}_c\delta_R}{\sum_{z_j\in c}Q_j} + \frac{n_c^2\bar{\Phi}_{c}Q_{i}\delta_R}{(\sum_{z_j\in c}Q_{j})^2},
\end{align}

$\bar{\Phi}_c$ and $Q_i$ cannot be arbitrarily large as they are the average Shapely value for a cluster and change in performance due to a data point $z_i$. With increasing cluster size $n_c$, the corresponding predicted value $\sum_{z_j\in c}Q_j$ will increase. Thus, $n_c / \sum_{z_j\in c}Q_j$ can not be very large. The $\delta_R$ error will be low for a moderately good regression model. Thus, our method estimates the Shapely value $\Phi_i$ with negligible error.

\subsection{Corollary 4.3.1}
The value $V_c$ of the a cluster by Shapley axioms is defined as
\begin{align}
V_c & = \sum_{z_j \in c} \Phi(z_j) = n_c\bar{\Phi}_c + \sum \delta(z_j).
\label{eq:cluster_value}
\end{align}

The sum of the Shapley values within a cluster equalling the leave-cluster-out value of the cluster follows logically from the axioms of the Shapley value, in particular, additivity and efficiency.

\textit{Additivity:} If we treat each data point as contributing a separate game to the performance, the total Shapley value of a cluster should naturally be the sum of the Shapley values of each data point within the cluster.\par

\textit{Efficiency:} This axiom ensures that the total value generated by the coalition is fully distributed among the players. If the Shapley value calculation respects this axiom, then the allocation to a cluster should match the cumulative contribution of its members.\par

We can consider a simple proof by induction on the number of data points in the cluster:
\\\textbf{1. Base Case:} If a cluster has only one data point, then the Shapley value of the cluster is clearly equal to the Shapley value of the single data point.
\\\textbf{2. Inductive Step:} Assume the proposition holds for all clusters of size $k$. For a cluster of size $k+1$, if you remove one data point, by the induction hypothesis, the sum of the Shapley values of the remaining $k$ data points equals the value of these $k$ data points. Adding the $k+1$-st point, by the additivity axiom, the overall Shapley value would be the sum of the individual Shapley values.



\appendix
\end{document}